\documentclass[lettersize,journal]{IEEEtran}
\usepackage{amsmath,amsfonts}
\usepackage{algorithmic}
\usepackage{algorithm}
\usepackage{array}
\usepackage[caption=false,font=footnotesize]{subfig}
\usepackage{textcomp}
\usepackage{stfloats}
\usepackage{url}
\usepackage{verbatim}
\usepackage{graphicx,import}
\usepackage{cite}
\usepackage{nicefrac}
\usepackage{booktabs}
\usepackage{svg}
\usepackage{todonotes}
\usepackage{fancyhdr}
\fancypagestyle{firstpage}{
  \fancyhf{}
  
  \fancyfoot[C]{\fontsize{6.5}{8} \selectfont \textbf{Copyright Note:} \copyright~2023 IEEE. This is the author's version of an article that has been accepted for publication in IEEE Transactions on Medical Robotics and Bionics. Personal use of this material is permitted. Permission from IEEE must be obtained for all other uses, in any current or future media, including reprinting/republishing this material for advertising or promotional purposes, creating new collective works, for resale or redistribution to servers or lists, or reuse of any copyrighted component of this work in other works. DOI: 10.1109/TMRB.2024.3349626}
}

\begin{document}

\title{Towards Safe and Collaborative Robotic \\Ultrasound Tissue Scanning in Neurosurgery}
\author{Michael Dyck*, Alistair Weld*, Julian Klodmann, Alexander Kirst, Luke Dixon,\\Giulio Anichini, Sophie Camp, Alin Albu-Schäffer, and Stamatia Giannarou
\\
\textit{michael.dyck@dlr.de, a.weld20@imperial.ac.uk}

\thanks{Michael Dyck* and Alin Albu-Schäffer are with the Institute of Robotics and Mechatronics, German Aerospace Center (DLR), Germany; and with the TUM School of Computation, Information and Technology, Technical University of Munich, Germany.}
\thanks{Alistair Weld* and Stamatia Giannarou are with the Hamlyn Centre for Robotic Surgery, Imperial College London, UK.}
\thanks{Julian Klodmann and Alexander Kirst are with the Institute of Robotics and Mechatronics, German Aerospace Center (DLR), Germany.}
\thanks{Luke Dixon, Giulio Anichini, and Sophie Camp are with the Department of Neurosurgery, Charing Cross Hospital, Imperial College London, UK.}
\thanks{*These authors contributed equally to the work.}
\thanks{Manuscript received October 31, 2022; accepted December 05, 2023.}}



\maketitle
\thispagestyle{firstpage}

\begin{abstract}
Intraoperative ultrasound imaging is used to facilitate safe brain tumour resection. 
However, due to challenges with image interpretation and the physical scanning, this tool has yet to achieve widespread adoption in neurosurgery. 
In this paper, we introduce the components and workflow of a novel, versatile robotic platform for intraoperative ultrasound tissue scanning in neurosurgery. 
An RGB-D camera attached to the robotic arm allows for automatic object localisation with ArUco markers, and 3D surface reconstruction as a triangular mesh using the \emph{ImFusion Suite} software solution.
Impedance controlled guidance of the US probe along arbitrary surfaces, represented as a mesh, enables collaborative US scanning, i.e., autonomous, teleoperated and hands-on guided data acquisition. 
A preliminary experiment evaluates the suitability of the conceptual workflow and system components for probe landing on a custom-made soft-tissue phantom.
Further assessment in future experiments will be necessary to prove the effectiveness of the presented platform.
\end{abstract}


\section{Introduction}
\IEEEPARstart{B}{rain} tumours cause more deaths than any other cancer for adults under the age of 40 \cite{rouse2015}.
Complete resection of cancerous tissue is one of the main treatments for brain tumours \cite{nitta1995}. 
There is evidence that increasing the extent of tumour resection substantially improves overall and progression-free survival \cite{devaux1993}. 
To facilitate maximal safe brain tumour resection, real-time intraoperative imaging tools are vital for identification of residual pathological tissue and intraoperative replanning \cite{bucci2004}.
Intraoperative Ultrasound (iUS) is an established tool for tissue characterisation \cite{sastry2017}. 
Unlike alternative tools, which require an extensive disruption of the surgical workflow \cite{bastos2021}, ultrasound (US) can be used with minimal interruption to the workflow and provides real-time imaging. 
What restrains iUS from widespread adoption is the difficulty in capturing and interpreting the data. 
Obtaining high quality images requires experienced personnel and is a strenuous workload (both physically and cognitively) to the operator. 
Accuracy, reproducibility and standardisation are necessary to establish iUS in neurosurgery \cite{dixon2022}.
Robotic iUS scanning can assist neurosurgeons in capturing US data, reducing the workload and achieving high reproducibility - while helping with the shortage of sonographers and neuroradiologists.
\begin{figure}[t]
  \centering
  \fontsize{9}{9}
  \selectfont
  \includegraphics[width=0.8\linewidth]{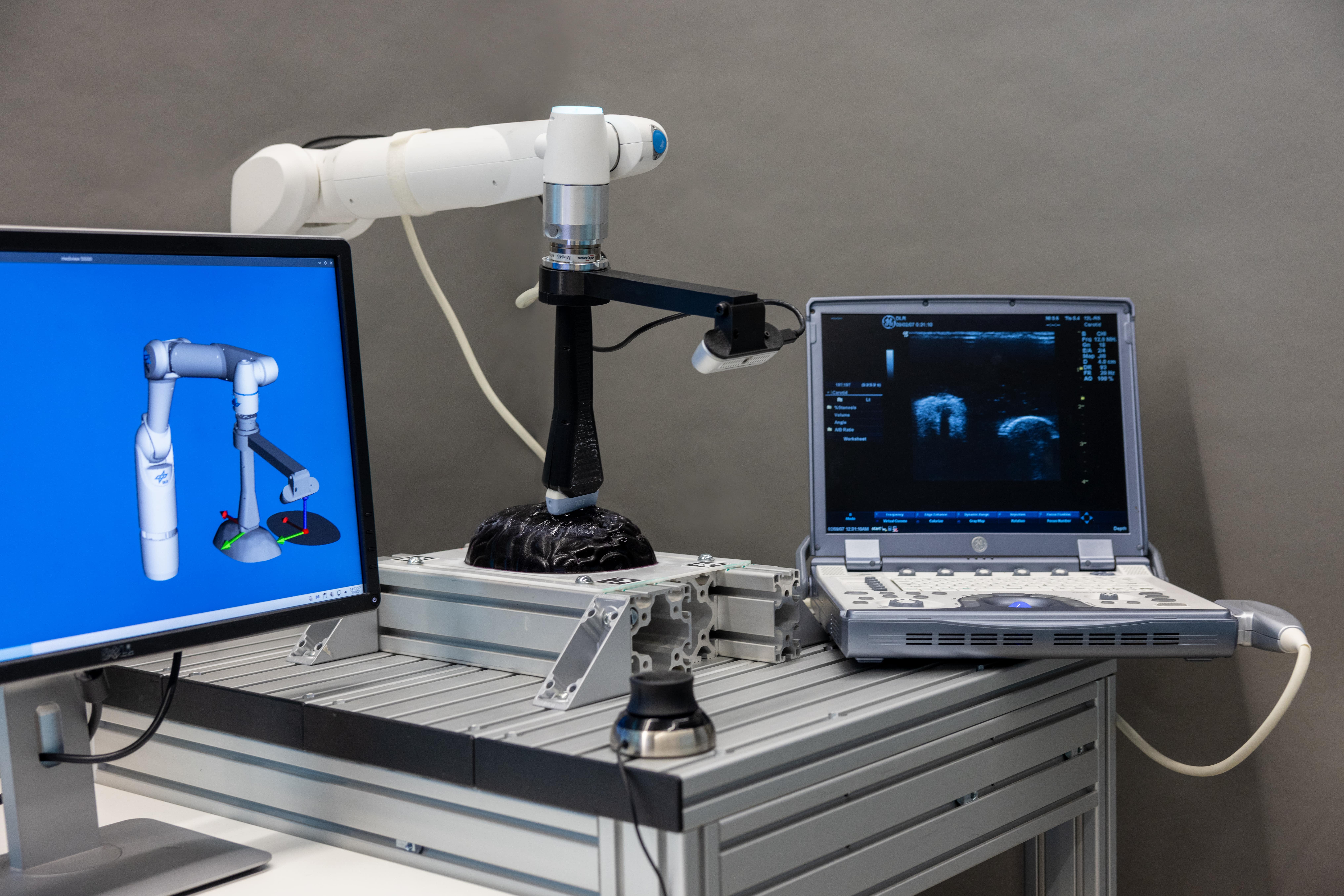}
  \caption{Experimental setup showing the robotic arm holding US probe and camera next to the US machine, scanning a brain phantom. A SpaceMouse\textsuperscript{\textregistered} can be used to telemanipulate the US probe. \tiny{\emph{DLR/Alexandra Beier (CC BY-NC-ND 3.0)}}}
  \label{fig:setup}
\end{figure}

\smallskip

Various research groups work on the development of robotic US systems (RUS) \cite{li2021}. 
Most solutions deploy a Cartesian impedance controller with direct force control along three-dimensional (3D) surfaces, as e.g., in \cite{jiang2020}. 
Many investigate different approaches to visual servoing based on US data for image quality optimisation \cite{chatelain2015}, interaction force adjustment \cite{jiang2021}, or automated US scanning of the thyroid \cite{zielke2022}. 
Usually, purely autonomous execution is considered, and external imaging and tracking systems are used for registration and surface reconstruction. To the best of our knowledge, none of these works focus on iUS tissue scanning in brain surgery. 
The delicacy of scanning brain tissue poses major challenges which are not common in other medical applications. 
The tissue's very low stiffness requires low interaction forces. 
Even probe guidance without tissue contact can be necessary, using a coat of saline to achieve acoustic coupling - leaving direct force control impracticable to apply.
Besides, purely autonomous robotic US data acquisition might not be desirable, making telemanipulation or hands-on guidance important alternatives.
Furthermore, robotic solutions requiring the presence of external imaging and tracking have a larger hardware footprint and require longer installation times for, e.g. preoperative calibration, and thus complicate their integration into the OR.

In this work we propose a robotic iUS scanning platform, presenting an exemplary medical workflow for autonomous, teleoperated or hands-on guided brain tissue scanning. 
A custom-made mechanical interface to US transducer and stereo camera alleviates the need for external tracking.
Deploying the impedance controller presented in \cite{dyck2022} enables guidance of an US probe in different control modes, achieving passive interaction dynamics not requiring direct force control. 
By incorporating the tissue surface directly into the real-time control loop, setup and integration of the platform is straightforward.

\section{Methods}
\begin{figure*}[!ht]
  \centering
  \fontsize{9}{9}\selectfont
  \def\svgwidth{1.\textwidth}
  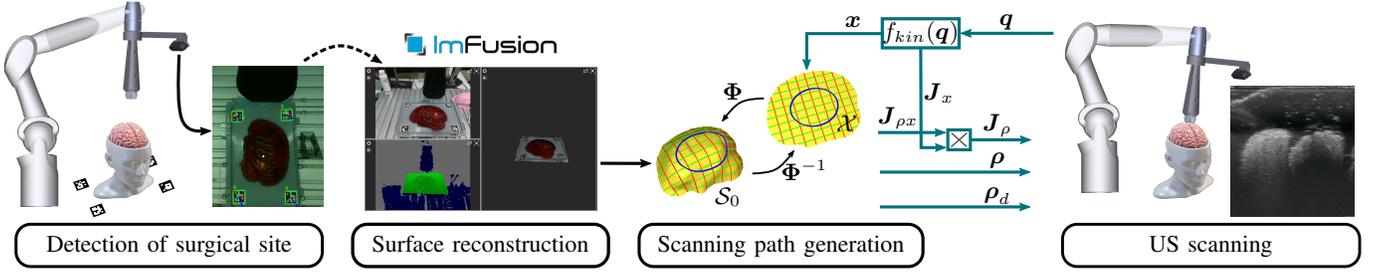
  \caption{Workflow for our iUS tissue scanning platform. (1) Automatic detection of the object's location using ArUco markers; (2) RGB-D reconstruction (with ImFusion) of phantom surface with a stereo camera results in a triangular mesh; (3) Incorporation of mesh geometry into the real-time control loop; (4) US probe guidance for contact establishment and tissue scanning. Note that the upper half of the brain being fully exposed in this image is merely for the purpose of visualisation. In reality, the craniotomy will be much smaller.}
  \label{fig:workflow}
\end{figure*}

The workflow of the presented robotic iUS tissue scanning platform is depicted in Fig.~\ref{fig:workflow}. This section introduces the individual components and their role within the system.

\subsection{Hardware Integration} \label{sec:hardware}
\begin{figure}[t]
	\centering
	\fontsize{9}{9}
  	\selectfont
    \subfloat[Mechanical tool interface.]{
    	\label{fig:transformations}
    	\fontsize{9}{9}\selectfont
  		\def\svgwidth{0.65\columnwidth}
  		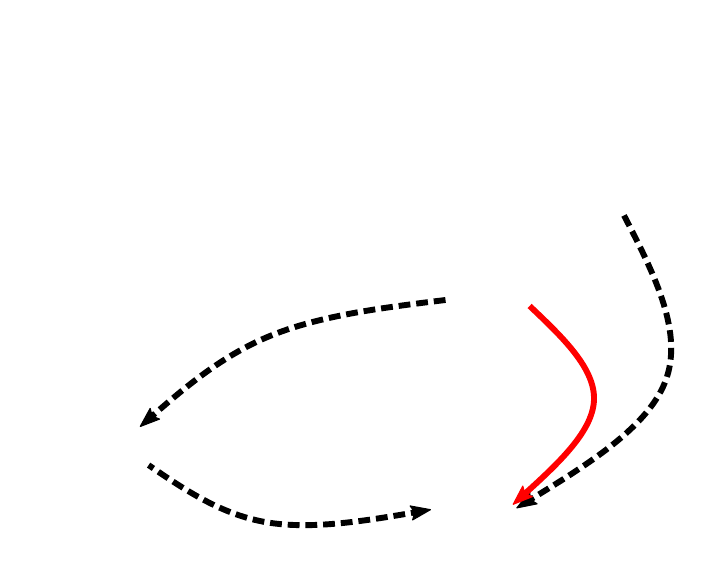}
    \subfloat[Brain phantom.]{
    	\label{fig:phantom}
    	\fontsize{9}{9}\selectfont
  		\def\svgwidth{0.3\columnwidth}
  		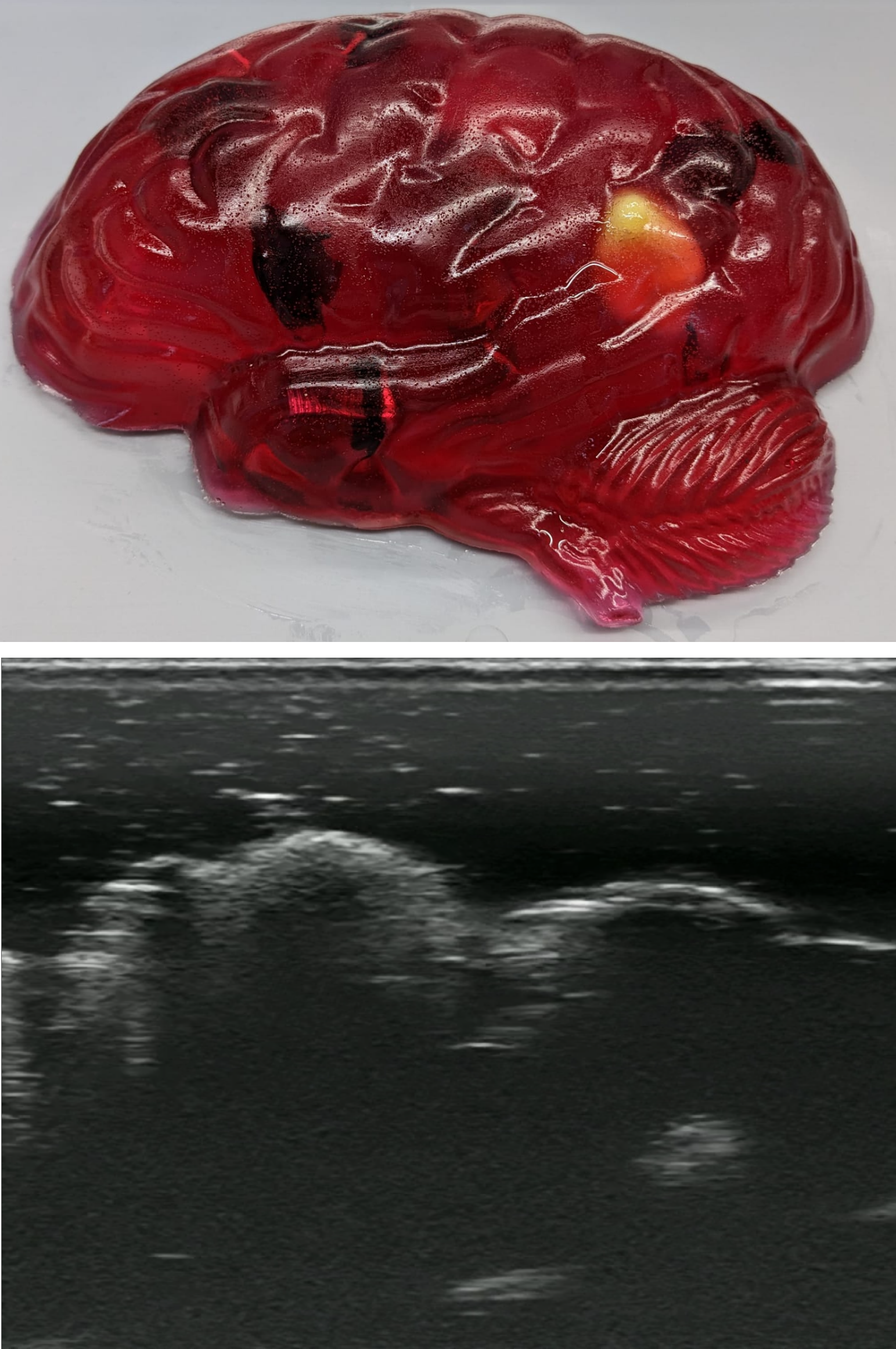}
	\caption{(a) Frames: $\{\mathcal{B}\}$ robot base, $\{\mathcal{C}\}$ camera, $\{\mathcal{P}\}$ US probe, $\{\mathcal{O}\}$ object. (b) Custom-made brain phantom with sample US image.}
	\label{fig:hardware}
\end{figure}

We designed a 3D printable mechanical interface between robot, camera and US probe. 
Fig.~\ref{fig:transformations} shows a model of a robotic arm with the tool interface attached to its end-effector.
Here, \begin{equation}
    T_{\mathcal{X}\mathcal{Y}}=
    \begin{bmatrix}
        R_{\mathcal{X}\mathcal{Y}} & p_{\mathcal{X}\mathcal{Y}}\\
        \boldsymbol{0}_{1\times3} & 1
    \end{bmatrix}
    \in SE(3)
\end{equation}
is a homogeneous transformation of frame $\{\mathcal{Y}\}$ with respect to frame $\{\mathcal{X}\}$. 
$R_{\mathcal{X}\mathcal{Y}} \in SO(3)$ is the corresponding rotation matrix, $p_{\mathcal{X}\mathcal{Y}} \in \mathbb{R}^3$ is the position vector between both frames.
Performing an accurate hand-eye calibration following \cite{strobl2006optimal}, and computing the forward kinematics mapping provides the camera's intrinsic parameters and the homogeneous transformation matrix $T_{\mathcal{B}\mathcal{C}}$ (base frame $\{\mathcal{B}\}$, camera frame $\{\mathcal{C}\}$).
The transformation $T_{\mathcal{B}\mathcal{P}}$ (US probe frame $\{\mathcal{P}\}$) originates from the CAD model of the tool interface and can be used without the need of an external tracking system. As we do not extract features from the US images in this work, but merely control the tip of the US transducer, no additional hand-eye calibration is necessary for this sensor.

Given this, we calculate $T_{\mathcal{P}\mathcal{O}}$ using the relation (see Fig.~\ref{fig:transformations}) \begin{equation}
    T_{\mathcal{P}\mathcal{O}}=T_{\mathcal{B}\mathcal{P}}^{-1}T_{\mathcal{B}\mathcal{O}}=T_{\mathcal{B}\mathcal{P}}^{-1}T_{\mathcal{B}\mathcal{C}}T_{\mathcal{C}\mathcal{O}}.
\end{equation}
Identification of the object pose relative to the camera $T_{\mathcal{C}\mathcal{O}}$ is explained in Sec.~\ref{sec:reconstruction}.

\subsection{Soft-Tissue-Mimicking Brain Phantom} \label{sec:phantom}
The custom-made soft-tissue phantom depicted in Fig.~\ref{fig:phantom} uses GELITA\textsuperscript{\textregistered} GELATINE \textit{Type Ballistic 3} as basis. To mimic the properties of brain tissue and to increase durability, the phantom contains water, glycerine, and gelatine at a ratio of 45:45:10. Inspired by \cite{Morehouse2007AdditionOM}, we placed different objects arbitrarily inside the phantom, including olives, grapes, blueberries, and screws. This ensures the presence of various features within the US recordings of the phantom, replicating the inhomogeneity of real tissue.

\subsection{Localisation and Surface Reconstruction} \label{sec:reconstruction}
For localisation of the phantom within the robot's workspace, four ArUco markers are first manually placed around the object (cmp. Figs.~\ref{fig:setup},\ref{fig:workflow}). Positioning the RGB-D camera above the area of interest, such that all markers are visible within the image, initialises the automatic localisation and reconstruction routine. Using the OpenCV \cite{itseez2015opencv} ArUco library, the four fiducial markers are identified. Based on these points, a plane - defined by a centre point and normal vector - containing the phantom is calculated. The robot automatically aligns the camera such that it's z-axis points towards the plane's centre at an angle of $45\,^{\circ}$ and a distance of $30$\,cm. The RGB-D camera is then rotated around the centre point/normal - capturing the phantom at multiple views. Using the image and depth stream, the RGB-D reconstruction algorithm from the \textit{ImFusion Suite} software solution by ImFusion GmbH reconstructs a 3D triangular mesh of the phantom surface. In doing so, we obtain object geometry and pose ($T_{\mathcal{C}\mathcal{O}}$).

\subsection{Collaborative Robotic Tissue Scanning}\label{sec:control}
The impedance controller introduced in previous work by Dyck et al. \cite{dyck2022} enables collaborative robotic US scanning. 
The recovered 3D morphological structure (cmp. Sec.~\ref{sec:reconstruction}), representing the anatomical surface as a triangular mesh, is utilised to define surface-specific coordinates. 
These coordinates are the distance $d\in\mathbb{R}$ between transducer and surface, as well as three orientational coordinates $\boldsymbol{\epsilon}\in\mathbb{R}^{3\times1}$, aligning the US probe axis with the surface normal. 
Two additional two-dimensional (2D) coordinates $(s_1, s_2)\in\mathcal{X}\subset\mathbb{R}^2$ describe the planar representation of the anatomical surface.
Stacking all these coordinates in a vector $\boldsymbol{\rho}=(s_1, s_2, d, \epsilon_1, \epsilon_2, \epsilon_3)^T\in\mathbb{R}^{6\times1}$, the joint torques are calculated according to the unified impedance control framework \cite{schaeffer2007}\begin{equation}\label{eq:imp}
	\boldsymbol{\tau} = \boldsymbol{J}_{\rho}^T(\boldsymbol{q})[\boldsymbol{K}_{\rho}(\boldsymbol{\rho}_d - \boldsymbol{\rho}(\boldsymbol{q}))+\boldsymbol{D}_{\rho}(\boldsymbol{q})(\boldsymbol{\dot{\rho}}_d - \boldsymbol{\dot{\rho}})].
\end{equation}
Here, $\boldsymbol{q}\in\mathbb{R}^{7x1}$ are the generalised configuration coordinates of the robot, $(\boldsymbol{\rho}_d,\boldsymbol{\dot{\rho}}_d)$ are the desired position and velocity, $\boldsymbol{J}_{\rho}\in\mathbb{R}^{6\times 6}$ represents the Jacobian matrix mapping joint velocities $\boldsymbol{\dot{q}}$ to velocities $\boldsymbol{\dot{\rho}}$. $\boldsymbol{K}_{\rho}$ and $\boldsymbol{D}_{\rho}(\boldsymbol{q})$ are the positive-definite and symmetric stiffness and damping matrices, respectively. 
Controlling the coordinate $d$ allows to realise different scenarios of probe-tissue interaction, such as (1) $d>0$: contact avoidance with a safety margin for, e.g., contact-free US scanning, (2) $d=0$: contact establishment, or (3) $d<0$: adjustment of the penetration depth to improve imaging quality. 
The capability of the controller to continuously adjust the penetration depth and interaction dynamics constitutes a unique feature, especially relevant for scanning of inhomogeneous, low stiffness tissue.
Passivity and stability are guaranteed by the implicit properties of impedance control during interaction and on contact loss.

\begin{figure}[t]
	\centering
	\fontsize{9}{9}
  	\selectfont
    \subfloat[Teleoperation.]{
    	\label{fig:teleoperation}
    	\fontsize{9}{9}\selectfont
  		\def\svgwidth{0.55\columnwidth}
\begingroup%
  \makeatletter%
  \providecommand\color[2][]{%
    \errmessage{(Inkscape) Color is used for the text in Inkscape, but the package 'color.sty' is not loaded}%
    \renewcommand\color[2][]{}%
  }%
  \providecommand\transparent[1]{%
    \errmessage{(Inkscape) Transparency is used (non-zero) for the text in Inkscape, but the package 'transparent.sty' is not loaded}%
    \renewcommand\transparent[1]{}%
  }%
  \providecommand\rotatebox[2]{#2}%
  \newcommand*\fsize{\dimexpr\f@size pt\relax}%
  \newcommand*\lineheight[1]{\fontsize{\fsize}{#1\fsize}\selectfont}%
  \ifx\svgwidth\undefined%
    \setlength{\unitlength}{156.69720795bp}%
    \ifx\svgscale\undefined%
      \relax%
    \else%
      \setlength{\unitlength}{\unitlength * \real{\svgscale}}%
    \fi%
  \else%
    \setlength{\unitlength}{\svgwidth}%
  \fi%
  \global\let\svgwidth\undefined%
  \global\let\svgscale\undefined%
  \makeatother%
  \begin{picture}(1,0.70871801)%
    \lineheight{1}%
    \setlength\tabcolsep{0pt}%
    \put(0,0){\includegraphics[width=\unitlength,page=1]{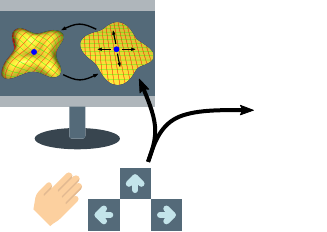}}%
    \put(0.78017435,0.26962952){\color[rgb]{0,0,0}\makebox(0,0)[t]{\lineheight{1.25}\smash{\begin{tabular}[t]{c}$\boldsymbol{\rho}_d(t)=\begin{bmatrix} s_{1,d}(t)\\s_{2,d}(t)\\d_d\\\boldsymbol{\epsilon}_d \end{bmatrix}$\end{tabular}}}}%
    \put(0,0){\includegraphics[width=\unitlength,page=2]{teleoperation2_svg-tex.pdf}}%
  \end{picture}%
\endgroup%
}
    \subfloat[Hands-on guidance.]{
    	\label{fig:handson}
    	\fontsize{9}{9}\selectfont
  		\def\svgwidth{0.35\columnwidth}
\begingroup%
  \makeatletter%
  \providecommand\color[2][]{%
    \errmessage{(Inkscape) Color is used for the text in Inkscape, but the package 'color.sty' is not loaded}%
    \renewcommand\color[2][]{}%
  }%
  \providecommand\transparent[1]{%
    \errmessage{(Inkscape) Transparency is used (non-zero) for the text in Inkscape, but the package 'transparent.sty' is not loaded}%
    \renewcommand\transparent[1]{}%
  }%
  \providecommand\rotatebox[2]{#2}%
  \newcommand*\fsize{\dimexpr\f@size pt\relax}%
  \newcommand*\lineheight[1]{\fontsize{\fsize}{#1\fsize}\selectfont}%
  \ifx\svgwidth\undefined%
    \setlength{\unitlength}{95.65288364bp}%
    \ifx\svgscale\undefined%
      \relax%
    \else%
      \setlength{\unitlength}{\unitlength * \real{\svgscale}}%
    \fi%
  \else%
    \setlength{\unitlength}{\svgwidth}%
  \fi%
  \global\let\svgwidth\undefined%
  \global\let\svgscale\undefined%
  \makeatother%
  \begin{picture}(1,0.74380402)%
    \lineheight{1}%
    \setlength\tabcolsep{0pt}%
    \put(0,0){\includegraphics[width=\unitlength,page=1]{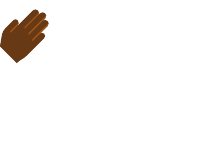}}%
    \put(0.77380295,0.38608745){\color[rgb]{0,0,0}\makebox(0,0)[t]{\lineheight{1.25}\smash{\begin{tabular}[t]{c}$\boldsymbol{p}_d=\begin{bmatrix} d_d\\\boldsymbol{\epsilon}_d \end{bmatrix}$\end{tabular}}}}%
    \put(0,0){\includegraphics[width=\unitlength,page=2]{handson2_svg-tex.pdf}}%
  \end{picture}%
\endgroup%
}
	\caption{Collaborative robotic tissue scanning controller.}
	\label{fig:controlmodes}
\end{figure}

Scanning trajectories along the surface can be planned in the 2D parameter domain $\mathcal{X}$ and automatically executed. 
As shown in Fig.~\ref{fig:controlmodes}, if desired, teleoperated and hands-on guided US scanning can be used instead.
Teleoperation opens the possibility for the scan to be performed by a remote expert.
The operator can use any (at least) 2D input device, to modify the desired position $(s_{1,d},s_{2,d})$ of the probe on the tissue (cmp. Fig.~\ref{fig:teleoperation}). 
A monitor displaying, both the original surface and its parametrised representation facilitates intuitive teleoperation.
Utilising only the coordinates $p = (d, \boldsymbol{\epsilon})\in\mathbb{R}^{4\times 1}$ and setting the impedance in $(s_1,s_2)$ to zero, allows the clinician to guide the US transducer along the surface (see Fig.~\ref{fig:handson}), to, e.g., spontaneously revisit certain anatomical structures during surgery.
In both cases, the controller is taking care of maintaining the specified distance and orientation.

\subsection{Workflow}

The workflow depicted in Fig.~\ref{fig:workflow} can be summarised as follows.
A manual, rough initialisation positions the end-effector above the object of interest. 
Following the object localisation using ArUco markers, the robot executes a rotational motion around the transducer axis, while the stereo camera records images of the target tissue surface and the RGB-D reconstruction algorithm from the \textit{ImFusion Suite} performs the 3D surface reconstruction (see Sec.~\ref{sec:reconstruction}). 
The resulting mesh and its pose relative to the robot are passed to the controller to calculate the set of surface-specific coordinates (see Sec.~\ref{sec:control}). 
The forward kinematics function $\boldsymbol{x} = f_{kin}(\boldsymbol{q})$ provides Cartesian coordinates $\boldsymbol{x}\in\mathbb{R}^{6\times 1}$ of the end effector, together with the geometrical Jacobian $\boldsymbol{J}_x\in\mathbb{R}^{6\times 7}$.
Left-multiplication of $\boldsymbol{J}_x$ with the Jacobian $\boldsymbol{J}_{\rho x}\in\mathbb{R}^{6x6}$ derived in \cite{dyck2022} gives $\boldsymbol{J}_{\rho}$ required in (\ref{eq:imp}).
The robot is landing the US transducer onto the tissue surface and initiates scanning trajectory execution with passive interaction dynamics, providing the option to change from autonomous execution to teleoperated, or hands-on probe guidance.

\section{Experimental Setup and Results}
The experimental setup can be seen in Fig.~\ref{fig:setup}. We use the DLR \textsc{Miro} surgical robotic arm \cite{seibold2018}, a seven degrees of freedom (DoF) robot arm developed specifically for medical applications. 
Redundant sensing results in high accuracy and safety, while the internal torque sensors provide an immediate interface to directly command joint torques. 
Attached is an Intel\textsuperscript{\textregistered} RealSense\texttrademark\ \textit{Depth Camera D435i}, and a \textit{GE 12L-RS} linear US probe.
The US transducer is connected to a \textit{GE LOGIQ e} US machine. An \textit{ATI Mini45} F/T-sensor measures six DoF interaction forces during scanning.
The robot is controlled at a frequency of $3$\,kHz, with the impedance controller (\ref{eq:imp}) running on a realtime computing host.

\begin{figure}[t]
  \centering
  \fontsize{9}{9}\selectfont
  \def\svgwidth{0.95\columnwidth}
  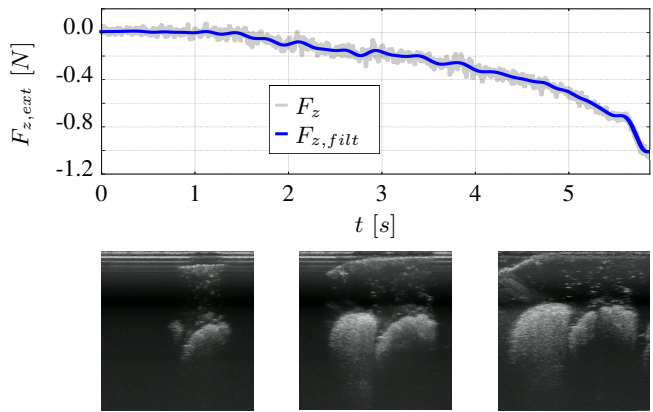
  \caption{Interaction force between US transducer and brain phantom during contact establishment. The US images below were recorded during the interaction and show qualitatively different stages of US contact.}
  \label{fig:contact_establishment}
\end{figure}

We perform one experiment to validate the presented workflow, and evaluate the suitability of its individual components for robot-guided probe landing on a soft-tissue phantom.
Qualtitative results of the automatic marker detection and 3D surface reconstruction can be seen in Fig.~\ref{fig:workflow}.
Fig.~\ref{fig:contact_establishment} depicts the force measured between the US probe and phantom during robot-guided contact establishment. 
The US probe is aligned normal to the surface ($\boldsymbol{\varepsilon}_d=\boldsymbol{0}_{3\times1}$), the setpoint for the distance coordinate ($d_d$) continuously decreases (increasing penetration depth). 
As expected, Fig.~\ref{fig:contact_establishment} shows that the force along the probe axis is continuously increasing. 
The US images in Fig.~\ref{fig:contact_establishment} were recorded during the experiment and show different stages of contact establishment - ranging from loose contact on the left to improved probe-tissue coupling on the right. 
These results are of merely qualitative nature, as interaction force and image quality highly depend on various factors, such as controller and object stiffnesses, and echogenic characteristics of the phantom. 
Nonetheless, this first preliminary experiment indicates the suitability of the proposed workflow and control approach for robotic probe landing. Particularly, the continuous adjustment of the penetration depth (and interaction dynamics) can be used to achieve different stages of probe-tissue coupling at low interaction force ($<1$\,N).

\section{Discussion and Conclusion}
This paper presented initial results from a novel robotic platform for iUS tissue scanning. 
The system integrates object localisation and reconstruction with novel robot control components, intended for application in neurosurgery for brain tumour resection guidance.
Relying on surface representation as triangular meshes, the deployed impedance controller enables easy integration of the imaging component.
A first preliminary experiment demonstrates the possibility to continuously adjust the penetration depth by controlling the distance between probe and tissue, indicating its suitability for iUS scanning of soft brain tissue.
In contrast to other systems, high performance force control at low forces and the challenging choice of the desired force profile to establish desired probe-tissue coupling is not required for the proposed controller.
Furthermore, the control concept makes the system collaborative, showing in its theoretical capability to execute autonomous, teleoperated or hands-on tissue scanning.
However, the significance of the experimental validation remains limited, as it does not evaluate the system's capability of maintaining certain interaction dynamics while scanning along the tissue.
To prove the effectiveness of the presented methods, further assessment and additional experiments will be necessary.

Future work will involve research towards a visual servoing concept, to control the US probe in a way that optimises data quality, by integrating a novel solution for iUS quality assessment \cite{weld2023identifying}. 
Furthermore, we are planning to enhance our platform to be capable of autonomously guiding the US probe to provide optimised intraoperative tissue characterisation, assisting surgeons in achieving maximal safe tumour resection.

\section*{Acknowledgments}
This project was supported by the International Graduate School of Science and Engineering (IGSSE); TUM Graduate School, and the UK Research and Innovation (UKRI) Centre for Doctoral Training in AI for Healthcare (EP/S023283/1), the Royal Society (URF$\setminus$R$\setminus$2 01014]), the NIHR Imperial Biomedical Research Centre. The authors want to thank \textit{ImFusion GmbH}, and especially Dr. Marco Esposito, for providing us an academic license to their software and constant support throughout our research.

\bibliographystyle{IEEEtran}
\bibliography{bibliography}

\begin{thebibliography}{10}
\providecommand{\url}[1]{#1}
\csname url@samestyle\endcsname
\providecommand{\newblock}{\relax}
\providecommand{\bibinfo}[2]{#2}
\providecommand{\BIBentrySTDinterwordspacing}{\spaceskip=0pt\relax}
\providecommand{\BIBentryALTinterwordstretchfactor}{4}
\providecommand{\BIBentryALTinterwordspacing}{\spaceskip=\fontdimen2\font plus
\BIBentryALTinterwordstretchfactor\fontdimen3\font minus
  \fontdimen4\font\relax}
\providecommand{\BIBforeignlanguage}[2]{{%
\expandafter\ifx\csname l@#1\endcsname\relax
\typeout{** WARNING: IEEEtran.bst: No hyphenation pattern has been}%
\typeout{** loaded for the language `#1'. Using the pattern for}%
\typeout{** the default language instead.}%
\else
\language=\csname l@#1\endcsname
\fi
#2}}
\providecommand{\BIBdecl}{\relax}
\BIBdecl

\bibitem{rouse2015}
C.~Rouse, H.~Gittleman, Q.~T. Ostrom, C.~Kruchko, and J.~S. Barnholtz-Sloan,
  ``Years of potential life lost for brain and cns tumors relative to other
  cancers in adults in the united states, 2010,'' \emph{Neuro-oncology},
  vol.~18, no.~1, pp. 70--77, 2015.

\bibitem{nitta1995}
T.~Nitta and K.~Sato, ``Prognostic implications of the extent of surgical
  resection in patients with intracranial malignant gliomas,'' \emph{Cancer},
  vol.~75, no.~11, pp. 2727--2731, 1995.

\bibitem{devaux1993}
B.~C. Devaux, J.~R. O'Fallon, and P.~J. Kelly, ``Resection, biopsy, and
  survival in malignant glial neoplasms: a retrospective study of clinical
  parameters, therapy, and outcome,'' \emph{Journal of neurosurgery}, vol.~78,
  no.~5, pp. 767--775, 1993.

\bibitem{bucci2004}
M.~K. Bucci, A.~Maity, A.~J. Janss, J.~B. Belasco, M.~J. Fisher, Z.~A. Tochner,
  L.~Rorke, L.~N. Sutton, P.~C. Phillips, and H.-K.~G. Shu, ``Near complete
  surgical resection predicts a favorable outcome in pediatric patients with
  nonbrainstem, malignant gliomas: results from a single center in the magnetic
  resonance imaging era,'' \emph{Cancer}, vol. 101, no.~4, pp. 817--824, 2004.

\bibitem{sastry2017}
R.~Sastry, W.~L. Bi, S.~Pieper, S.~Frisken, T.~Kapur, W.~Wells~III, and A.~J.
  Golby, ``Applications of ultrasound in the resection of brain tumors,''
  \emph{Journal of Neuroimaging}, vol.~27, no.~1, pp. 5--15, 2017.

\bibitem{bastos2021}
D.~C.~A. Bastos, P.~Juvekar, Y.~Tie, N.~Jowkar, S.~D. Pieper, W.~M. Wells,
  W.~L. Bi, A.~J. Golby, S.~F. Frisken, and T.~Kapur, ``Challenges and
  opportunities of intraoperative 3d ultrasound with neuronavigation in
  relation to intraoperative mri,'' \emph{Frontiers in Oncology}, vol.~11,
  2021.

\bibitem{dixon2022}
L.~Dixon, A.~Lim, M.~Grech-Sollars, D.~Nandi, and S.~Camp, ``Intraoperative
  ultrasound in brain tumor surgery: A review and implementation guide,''
  \emph{Neurosurgical Review}, pp. 1--13, 2022.

\bibitem{li2021}
K.~Li, Y.~Xu, and M.~Q.-H. Meng, ``An overview of systems and techniques for
  autonomous robotic ultrasound acquisitions,'' \emph{IEEE Transactions on
  Medical Robotics and Bionics}, vol.~3, no.~2, pp. 510--524, 2021.

\bibitem{jiang2020}
Z.~Jiang, M.~Grimm, M.~Zhou, Y.~Hu, J.~Esteban, and N.~Navab, ``Automatic
  force-based probe positioning for precise robotic ultrasound acquisition,''
  \emph{IEEE Transactions on Industrial Electronics}, vol.~68, no.~11, pp.
  11\,200--11\,211, 2020.

\bibitem{chatelain2015}
P.~Chatelain, A.~Krupa, and N.~Navab, ``Optimization of ultrasound image
  quality via visual servoing,'' in \emph{2015 IEEE International Conference on
  Robotics and Automation (ICRA)}, 2015, pp. 5997--6002.

\bibitem{jiang2021}
Z.~Jiang, Y.~Zhou, Y.~Bi, M.~Zhou, T.~Wendler, and N.~Navab,
  ``Deformation-aware robotic 3d ultrasound,'' \emph{IEEE Robotics and
  Automation Letters}, vol.~PP, pp. 1--1, 2021.

\bibitem{zielke2022}
J.~Zielke, C.~Eilers, B.~Busam, W.~Weber, N.~Navab, and T.~Wendler, ``Rsv:
  Robotic sonography for thyroid volumetry,'' \emph{IEEE Robotics and
  Automation Letters}, vol.~7, no.~2, pp. 3342--3348, 2022.

\bibitem{dyck2022}
M.~Dyck, A.~Sachtler, J.~Klodmann, and A.~Albu-Schäffer, ``Impedance control
  on arbitrary surfaces for ultrasound scanning using discrete differential
  geometry,'' \emph{IEEE Robotics and Automation Letters}, vol.~7, no.~3, pp.
  7738--7746, 2022.

\bibitem{strobl2006optimal}
K.~H. Strobl and G.~Hirzinger, ``Optimal hand-eye calibration,'' in \emph{2006
  IEEE/RSJ international conference on intelligent robots and systems}.\hskip
  1em plus 0.5em minus 0.4em\relax IEEE, 2006, pp. 4647--4653.

\bibitem{Morehouse2007AdditionOM}
H.~Morehouse, H.~P. Thaker, and C.~Persaud, ``Addition of metamucil to gelatin
  for a realistic breast biopsy phantom,'' \emph{Journal of Ultrasound in
  Medicine}, vol.~26, 2007.

\bibitem{itseez2015opencv}
Itseez, ``Open source computer vision library,''
  \url{https://github.com/itseez/opencv}, 2015.

\bibitem{schaeffer2007}
A.~Albu-Sch{\"a}ffer, C.~Ott, and G.~Hirzinger, ``A unified passivity-based
  control framework for position, torque and impedance control of flexible
  joint robots,'' \emph{The international journal of robotics research},
  vol.~26, no.~1, pp. 23--39, 2007.

\bibitem{seibold2018}
U.~Seibold, B.~K{\"u}bler, T.~Bahls, R.~Haslinger, and F.~Steidle, ``The {DLR
  MiroSurge} surgical robotic demonstrator,'' ser. The Encyclopedia of Medical
  Robotics, J.~P. Desai and R.~V. Patel, Eds.\hskip 1em plus 0.5em minus
  0.4em\relax WORLD SCIENTIFIC, October 2018, vol.~1, pp. 111--142.

\bibitem{weld2023identifying}
A.~Weld, L.~Dixon, G.~Anichini, M.~Dyck, A.~Ranne, S.~Camp, and S.~Giannarou,
  ``Identifying visible tissue in intraoperative ultrasound images during brain
  surgery: A method and application,'' 2023.

\end{thebibliography}

\end{document}